%% file: main.tex
\documentclass[10pt,twocolumn,letterpaper]{article}

\usepackage{wacv}
\usepackage{times}
\usepackage{epsfig}
\usepackage{graphicx}
\usepackage{amsmath}
\usepackage{amssymb}
\usepackage{booktabs}

\usepackage{microtype}      %
\usepackage{multirow}
\usepackage{makecell}

\usepackage{mathtools}
\usepackage[table]{xcolor}

\newcommand\ha{\rowcolor{orange!0}}

\newcommand*\samethanks[1][\value{footnote}]{\footnotemark[#1]}

\def\wacvPaperID{1896} %

\wacvalgorithmstrack   %

\wacvfinalcopy %

\ifwacvfinal
\usepackage[breaklinks=true,bookmarks=false]{hyperref}
\else
\usepackage[pagebackref=true,breaklinks=true,colorlinks,bookmarks=false]{hyperref}
\fi

\begin{document}

\title{Analysis of Quantization on MLP-based Vision Models}

\author{Lingran Zhao\thanks{Equal contribution.} \\
  Peking University \\
  {\tt\small calvinzhao@pku.edu.cn} \\
  \and
  Zhen Dong\samethanks,~ Kurt Keutzer \\
  University of California, Berkeley \\
  {\tt\small {zhendong,keutzer}@berkeley.edu} \\
}

\maketitle

\input{s0_abstract}
\input{s1_intro}
\input{s2_related_work}

\input{s3_methods}
\input{s4_experimental_results}
\input{s6_conclusion}

\clearpage
{\small
\bibliographystyle{ieee_fullname}
\bibliography{egbib}
}

\clearpage
\appendix
\input{s8_supp}

\end{document}

%% file: s0_abstract.tex
\begin{abstract}
Quantization is wildly taken as a model compression technique, which obtains efficient models by converting floating-point weights and activations in the neural network into lower-bit integers. Quantization has been proven to work well on convolutional neural networks and transformer-based models. Despite the decency of these models, recent works~\cite{touvron2021resmlp, fusco2022pnlp, ma2022rethinking} have shown that MLP-based models are able to achieve comparable results on various tasks ranging from computer vision, NLP to 3D point cloud, while achieving higher throughput due to the parallelism and network simplicity. However, as we show in the paper, directly applying quantization to MLP-based models will lead to significant accuracy degradation. Based on our analysis, two major issues account for the accuracy gap: 1) the range of activations in MLP-based models can be too large to quantize, and 2) specific components in the MLP-based models are sensitive to quantization. Consequently, we propose to 1) apply LayerNorm to control the quantization range of activations, 2) utilize bounded activation functions, 3) apply percentile quantization on activations, 4) use our improved module named multiple token-mixing MLPs, and 5) apply linear asymmetric quantizer for sensitive operations. Equipped with the abovementioned techniques, our Q-MLP models can achieve 79.68\% accuracy on ImageNet with 8-bit uniform quantization (model size 30 MB) and 78.47\% with 4-bit quantization (15 MB).
\end{abstract}

%% file: s1_intro.tex
\section{Introduction}
\label{sec:intro}

The deployment of Neural Network (NN) models is often impossible due to application-specific constraints on latency, power consumption, and memory footprint. 
This prohibits the use of state-of-the-art models with excellent accuracy but large parameter size and FLOPS. 
Quantization has been proposed as one of several model compression methods to enable efficient inference.
Generally, quantization converts floating-point NN models into integer-only or fixed-point models, which is efficient in memory consumption. Thanks to faster integer arithmetic compared to its floating-point counterparts, quantization can also reduce the computation when both weights and activations in the NN are quantized.

Since the success of traditional CNN networks~\cite{tan2019efficientnet, howard2019searching} and transformer-based networks~\cite{dosovitskiy2020image, liu2021swin} on various tasks from computer vision to NLP, significant research efforts have been spent on finding better building blocks for NN architectures. Recent works~\cite{tolstikhin2021mlp, touvron2021resmlp, liu2021pay, fusco2022pnlp, ma2022rethinking} have proposed that NN models based on Multi-Layer-Perceptron (MLPs) can also achieve state-of-the-art performance on those tasks. In addition to accuracy, MLP-based models benefit from their intrinsic parallelism and model simplicity and can potentially achieve higher throughput compared to CNNs and transformers. As Table \ref{tab:Throughput} shows, ResMLP has larger throughput than ViT and ResNets despite having more parameters and FLOPs.

\begin{table}[!htbp]
\caption{Comparing MLP-based models with transformers and CNNs. The throughput is measured on a single TITAN RTX 2080Ti (24GB) GPU with batch size fixed to 128. For reference, the accuracy included here is obtained by models trained solely on ImageNet with no extra data.}
\vspace{5pt}
\label{tab:Throughput}
  \begin{center}
    {\small{
\resizebox{0.99\linewidth}{!}{
\setlength\tabcolsep{2 pt}
\begin{tabular}{lccccc} \toprule
\multicolumn{1}{c}{\textbf{Model}} 
&\multicolumn{1}{c}{\textbf{Params }} 
&\multicolumn{1}{c}{\textbf{Throughput} }
&\multicolumn{1}{c}{\textbf{FLOPs} }
&\multicolumn{1}{c}{\textbf{Top-1} }
\\
 &($\times 10^6$) &(img/sec) &(G) &(\%) \\
\midrule
\ha 
ResMLP-S24/16 \cite{touvron2021resmlp} 
& 30 &468     & 11.94      &79.4    \\ 
ResMLP-B24/16 \cite{touvron2021resmlp}
& 115 &195     & 46.08    &81.0   \\
\midrule
\ha ViT-S/16 \cite{dosovitskiy2020image}
& 22 &451     & 8.48    &78.1   \\
ViT-B/16 \cite{dosovitskiy2020image}
& 86 &255     & 33.72    &79.9   \\
\midrule
\ha ResNet-50 \cite{he2016deep}
&26 &466     &7.76     &77.7   \\
ResNet-101 \cite{he2016deep}
&45 &287     &15.20    &79.2   \\
\bottomrule 
\end{tabular}
}}
}
\end{center}
\end{table}

Specifically, each block in MLP-based models has the same parameter size and the same resolution of feature maps,  whereas the blocks at the beginning of CNNs tend to have a much smaller parameter size and a larger resolution of feature maps than the subsequent blocks. This uniformity of building blocks makes MLP-based models easier to deploy and optimize on the hardware platforms compared to CNNs. Furthermore, uniform blocks are also friendly to uniform quantization. In contrast, when applying ultra-low bit quantization on CNNs, mixed-precision quantization~\cite{wang2018haq, dong2019hawq, dong2019hawqv2} is often required to alleviate the accuracy degradation, for which the hardware support can be sub-optimal. Compared to transformers, MLP-based models are also more efficient since they can avoid intensive computation~\cite{tolstikhin2021mlp} by not explicitly applying the attention mechanism.

In order to simultaneously achieve high accuracy and efficient inference, it is natural to explore quantization on MLP-based models. However, directly applying quantization to MLP-based models will lead to high accuracy degradation. In this work, we first find that the range of activations in specific MLP-based models can become too large to quantize. Consequently, we propose to restrict the activation range with carefully designed normalization and activation layers. From our experiments, applying LayerNorm instead of the Affine operation, utilizing bounded activation functions, and applying percentile quantization for activations proved beneficial in reducing the activation range. Secondly, our analysis shows that specific operations are more sensitive than the others in MLP-based models. To tackle this issue, we propose a new component named multiple token-mixer, which can be both efficient and less sensitive to quantization. Furthermore, applying asymmetric linear quantizers onto or after sensitive operations helps improve accuracy, with a trivial overhead to support the mixture of symmetric and asymmetric quantizers. 
Our contributions can be summarized as follows:
\begin{itemize}
    \item We are the first to analyze the causes of significant accuracy degradation when quantizing MLP-based models.
    \item We provide universal instructions for designing MLP-based models in order to make them more quantization-friendly.
    \item Our proposed quantization methods can achieve 79.68\% accuracy on ImageNet with 8-bit quantization (model size 30 MB), and our 4-bit quantized model has 78.47\% accuracy with only 15 MB model size.
\end{itemize}

%% file: s2_related_work.tex
\section{Related work}

\textbf{Quantization}~\cite{zhou2017incremental,jacob2018quantization,zhang_2018_lqnets,wang2018haq, cai2020zeroq, gholami2021survey} are common model compression techniques where low-bit precision is used for weights and activations to reduce model size without changing the original network architecture.
Quantization can also potentially permit the use of low-precision matrix multiplication or convolution, making the inference process faster and more efficient.

Despite these advances, directly performing post-training quantization (PTQ) with uniform ultra-low bit-width still results in a significant accuracy degradation. As such, Quantization-aware training (QAT) is proposed to train the model to better adapt to quantization. 
Another promising direction is to use mixed-precision quantization~\cite{zhou2017adaptive,wang2018haq,yao2021hawq}, where some layers are kept at higher precision.
Although mixed-precision quantization can be well supported on some existing hardware (such as FPGAs)~\cite{huang2021codenet,dong2021hao}, it can lead to a non-trivial overhead on many other hardware platforms (such as GPUs).

\textbf{MLP-based Models} have been recently proposed to perform various tasks, competing against previous convolutional neural networks (CNNs) and transformers. The MLP-Mixer~\cite{tolstikhin2021mlp} architecture, built entirely on multi-layer perceptrons (MLPs), has produced competitive results in vision tasks. Due to its simple and uniform structure, MLP-Mixer achieves high throughput and brings new possibilities to efficient-learning topics. 

Another important characteristic of MLP-Mixer is that 
it separately uses a channel-mixing MLP to enable communications between different feature channels within each token and a token-mixing MLP to enable communications between different spatial locations across patches. This two-step process in each layer increases the interpretability of deep neural networks and enables further investigating and special designing of each part in later works. In a subsequent work \cite{touvron2021resmlp}, the authors propose the architecture ResMLP, which simplifies the token-mixing module and the norm-layer, achieving a better efficiency-accuracy trade-off. 
Later, \cite{liu2021pay,yu2021rethinking,yu2022s2} further pushes the limits of MLP-based models by finding better ways to improve token-mixing and channel-mixing simultaneously.
A very recent work~\cite{trockman2022patches} combines the merits of convolutions and this mixer-based communication separating technique and proposes ConvMixer, which outperforms not only CNNs but also vision transformers and MLP-Mixer variants. It should be noted that, although ConvMixer is not precisely composed of MLPs, it is intrinsically similar to MLP-based models rather than CNNs. Therefore, we still conduct a detailed analysis of it due to its mixer-based structure and state-of-the-art performance.

%% file: s3_methods.tex
\section{Methodology}\label{sec:method}

In this section, we introduce a set of quantization techniques to combine the merit in MLP structure and the efficiency of quantization. We demonstrate that the MLP-based model provides inherent advantages for uniform quantization and can achieve a satisfying accuracy-efficiency trade-off when provided with appropriate techniques.

\subsection{Quantization preliminaries}
Quantization methods quantize weights and activations into integers with a scale factor $S$ and a zero point $Z_0$. Uniformly quantizing activations or weights to k bit can be expressed as:
\small
\begin{equation}
\begin{aligned}
S &=\frac{r_{\max }-r_{\min }}{2^{k}-1} \\
Z_{0} &=\operatorname{round}\left(2^{k-1}-1-\frac{r_{\max }}{S}\right) \\
q &=\operatorname{round}\left(\frac{r}{S}+Z_{0}\right)
\end{aligned}
\end{equation}
\normalsize
where $r$ is the real number and $q$ is the quantized integer and can then be used to enable hardware integer arithmetic acceleration. In baseline methods, we use symmetric quantization, which means $Z_0$ equals 0, and the first bit of $q$ serves as a sign bit, with the rest k-1 bits used to represent the integer. (More details in Section \ref{sec:sensitive})

There are usually two types of quantization methods: post-training quantization (PTQ) and quantization-aware training (QAT). For PTQ, we apply the above quantization directly in the inference stage to the pre-trained weights, and we use the quantized weights and activations to generate results. For QAT, we define the forward and backward pass for the above quantization operations and train quantized parameters together with the model parameters in order to get better quantization results. Both methods are useful and can be applied to different circumstances for deployment.

\subsection{MLP-based structures}
To make this paper self-contained, we briefly introduce the structure of MLP-based models here. More details can be found in \cite{tolstikhin2021mlp}.
MLP-based architectures work on image patches, and it uses two separate parts in each layer to enable communications between different patches and between different embedded channels. However, it is different from the Vision Transformer~\cite{dosovitskiy2020image} in that it uses only MLP modules to achieve these two goals, while Vision Transformer uses multi-head self-attention to enable communications between different patches. In the rest of this paper, we will see that this simplicity benefits MLP-based structures in uniform bitwidth quantization.

More concretely, MLP-Mixer has two MLPs in each layer. The first is a token-mixing MLP (note that in MLP-based models, patches are usually called tokens, and channels refer to the embedded patch features) that acts on each channel dimension with shared parameters. The second is a channel-mixing MLP which mixes the channels within each token. To sum up, MLP-Mixer layers can be written as follows:

\small
\begin{equation}
\begin{aligned}
Z_{*, i}=&X_{*, i}+W_{2} \sigma\left(W_{1} \operatorname{LayerNorm}(X)_{*, i}\right),\\
Y_{j, *}=&Z_{j, *}+W_{4} \sigma\left(W_{3} \operatorname{LayerNorm}(Z)_{j, *}\right),\\
\text { for } &i \in \{1, 2, \ldots C\}, j \in \{1, 2, \ldots S\}.   
\end{aligned}
\end{equation}
\normalsize

where ${X}$ refer to the input and has a shape of tokens ($S$) $\times$ channels ($C$), $\sigma$ is an element-wise nonlinearity, and $W_{i}$ refer to FC (Fully-Connected) weights in MLPs.

Subsequent MLP-based models have similar structures and only change in the ways of mixing. ResMLP proposes to use a simple linear transformation for token-mixing and replace LayerNorm with Affine for better efficiency. ConvMixer proposes to use convolutions instead of fully-connected layers to enable cross-token and cross-channel communications. Both of these two ideas have been influential in subsequent papers on MLP-based architecture design.

\begin{figure*}[t]
\centering
\includegraphics[width=.92\textwidth, trim=30 40 30 40, clip]{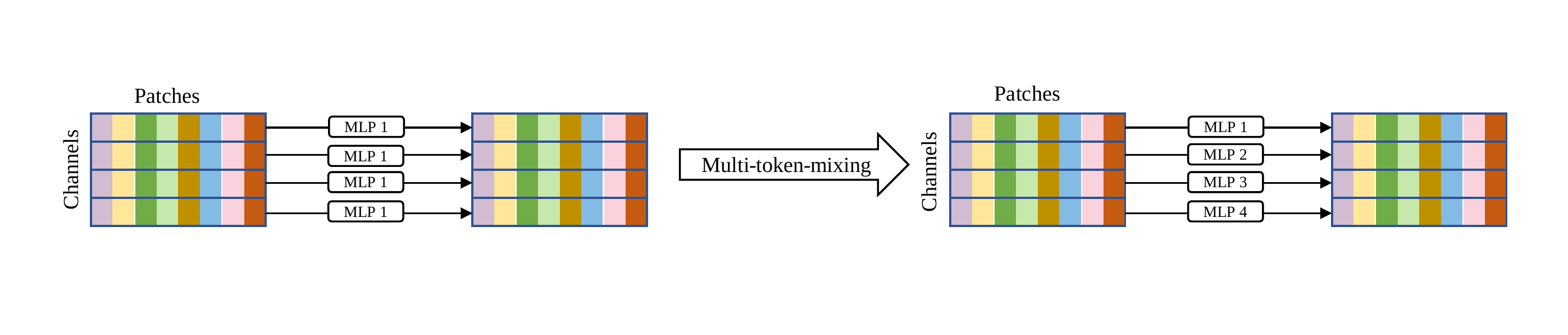}
\caption{
 Token-mixing module using multiple token-mixing MLPs.
}
  \label{fig:multitoken}
\end{figure*}

\subsection{Restrict activation ranges}
The magnitude of the activation range is highly related to the performance of the quantized models. Generally, a more extensive activation range loses more information with a given bitwidth quantization than a small activation range. In the experiments, we found that the activation ranges of some MLP-based models (e.g., ResMLP and ConvMixer) are unusually high, which leads to severe accuracy degradation in the quantized model. Therefore, it is of great importance to carefully deal with these activation ranges and use techniques to restrict them.

\paragraph{Norm-layer design}
The choice of norm layer significantly impacts the activation range of features and, therefore, is crucial to the PTQ performance of MLP-based models. Different MLP-based models use different norm layers, which lead to very different activation ranges.

Some models use a simple Affine transformation (Equation \ref{eq:aff}) as the norm-layer, which only rescales and shifts the input in an element-wise manner. Though it is demonstrated to be a slightly simpler and more efficient layer than LayerNorm, we found it potentially leads to a huge activation range (more details in Section~\ref{sec:restrict_activation_range}) and incurs an accuracy drop in its quantized model. 
\small
\begin{equation}\label{eq:aff}
\operatorname{Aff}_{\alpha, \beta}(x)=\mathrm{Diag}(\alpha) x+\beta
\end{equation}
\normalsize

Therefore, we proposed to replace Affine transformation with LayerNorm or BatchNorm (both can be represented by Equation \ref{eq:layernorm}) in all the MLP-based models in order to restrict the activation range using channel/batch statistics.
\small
\begin{equation}\label{eq:layernorm}
y=\frac{x-\mathrm{E}[x]}{\sqrt{\operatorname{Var}[x]+\epsilon}} * \gamma+\beta
\end{equation}
\normalsize

\paragraph{Activation layer design}
The choice of activation layer is another critical factor that affects the activation range in MLP models. Most MLP models use ReLU or GELU as activation layers, and they have been tested to have similar performance~\cite{touvron2021resmlp}. However, both GELU and ReLU are not the best choice for quantized MLP-based models since they are not bounded when activations are positive. We propose that the best activations for quantized MLP-based models are ones that are both bounded in negative input values and positive input values.

Parametrized clipping activation (PACT) (Equation \ref{eq:pact}), which we apply in our experiments, is one of the good choices for the activation layer of MLP-based models. It sets a learnable upper bound parameter to clip all the input values into the range of [0,$\alpha$]. (More details in paper~\cite{choi2018pact})
\small
\begin{equation}\label{eq:pact}
\begin{aligned}
y=P A C T(x)&=0.5(|x|-|x-\alpha|+\alpha) \\
&=\left\{\begin{array}{ll}
0, & x \in(-\infty, 0) \\
x, & x \in[0, \alpha) \\
\alpha, & x \in[\alpha,+\infty)
\end{array}\right.
\end{aligned}
\end{equation}
\normalsize
In practice, this kind of activation layer can largely restrict the activation range of the MLP-based model and lead to a better performance in quantized MLP models.

\paragraph{Percentiles in activation range}
Though some MLP-based models have a very large activation range, it does not necessarily imply that they have a large mean value of activations. Instead, the large range may be caused by some extreme outliers in the outputs. When this happens, we can significantly recover the performance by using percentiles to clip the extreme activation values.

\subsection{Tackle sensitive layers}\label{sec:sensitive}
Parameters and activations in different layers have relatively different sensitivity. Mixed-precision quantization methods allocate different bitwidth for different layers to overcome this problem. However, in the context of uniform quantization, we cannot tune a set of different bitwidths, so we proposed two alternative methods to tackle this issue.

\paragraph{Multiple token-mixing MLPs}
Usually, MLP-based models have two modules in each layer: the token-mixing module and the channel-mixing module. However, the two modules are different in parameter size and sensitivity. We use the Hessian trace analysis~\cite{yao2020pyhessian} to evaluate the sensitivity of the token-mixing MLPs and the channel-mixing MLPs in MLP-Mixer and present the results in Section \ref{sec:epsensitive} Figure \ref{fig:trace}. We found that the average sensitivity of the parameters (indicated by the mean Hessian trace of the learnable parameters) in token-mixing MLPs is much higher than that in channel-mixing MLPs. We can also comprehend this from a different perspective: since the channel dimensions are usually 4-5 times the token dimensions, parameters in token-mixing MLPs are usually reused 3-4 times more than that of channel-mixing MLPs (because the same MLP applies to all the token/channel dimensions). Therefore, the parameters in token-mixing MLPs are intuitively more sensitive to changes.

The above analysis explains why many subsequent papers performed better in MLP-based models after redesigning the token-mixing MLPs. It also indicates that we should carefully deal with parameters in token-mixing MLPs since performance drop in post-training quantization is highly related to parameter sensitivity.

Consequently, we modify the structure of the original token-mixing MLPs to reduce their sensitivity. As shown in Figure \ref{fig:multitoken}, different from applying the same token-mixing MLP to each of the C different channels in the original MLP-Mixer layers, we divide the channels into several groups and apply different token-mixing MLPs to different channel groups. This approach reduces the reuse of parameters in token-mixing MLPs and increases the expressibility of token MLPs. The experiments show that both the accuracy of MLP-Mixer and the accuracy of the post-training quantized MLP model increase after introducing the multiple token-mixing MLPs. Meanwhile, since the number of the parameters in channel-mixing MLPs is 30 times larger than that in token MLPs, using multiple token-mixing MLPs will not significantly increase the total parameter size of the model. Moreover, thanks to the merits of quantization, the model size of the quantized multiple token-mixing Mixer is still much smaller than the original MLP-Mixer.

\paragraph{Asymmetric quantization}
For MLP-based models, sensitivity imbalances are not only found in weights in different MLPs but also found between weights and activations. In the experiments, we find that activations are much more sensitive than weights in MLP-based models. This argument is derived from the fact that we can have a relatively good PTQ result with ultra-low bitwidth (3 or 4) weight quantization and 8-bit activation quantization, while we cannot get any acceptable PTQ results with an activation bit less than 8 (no matter how many bit the weights use).

To better deal with these sensitive activations in the context of uniform quantization, we propose to use asymmetric quantization for activations and still use symmetric quantization for weights. The term asymmetric quantization means that the zero point could be any floating point value, and the activation range depends on the difference in max and min of the input value instead of the max of the absolute value. More concretely, the scaling factor can be expressed as Equation \ref{eq:sym} for symmetric quantization:
\small
\begin{equation}\label{eq:sym}
S=\frac{\max(abs(r_{\max }),abs(r_{\min }))}{2^{k-1}-1}
\end{equation}
\normalsize
and Equation \ref{eq:asym} for asymmetric quantization,
\small
\begin{equation}\label{eq:asym}
S=\frac{r_{\max }-r_{\min }}{2^k-1}
\end{equation}
\normalsize
where r refers to the real number inputs and k refers to the activation bitwidth.
From Equation \ref{eq:sym} and \ref{eq:asym}, we can see that asymmetric quantization potentially provides one more bit for activation quantization (if the max positive and negative values are in different orders of magnitude). Therefore, we can better deal with sensitive activations while staying within the scope of uniform quantization.

%% file: s4_experimental_results.tex
\section{Experimental results}\label{sec:results}

To evaluate our proposed quantization approaches for MLP-based models, we perform a series of experiments on ImageNet with MLP-Mixer-B/16, ResMLP-S24/16, and ConvMixer-768/32 (since they have similar model scales). Results show that some of these techniques benefit the model’s accuracy, and others are even indispensable to avoid severe performance loss incurred by quantization.
We should note that, although ConvMixer does not have the MLP in its structure, it absorbs the idea in those recently proposed MLP-based models that the multi-head attention module can be replaced by MLP or convolution for higher efficiency. By default, we use uniform quantization throughout the layers to take advantage of the simplicity of MLP-based models, with weights and activations quantized to 8-bit for post-training quantization, and weights to 3/4/8 bit and activations to 8-bit for quantization-aware training. In addition, we use channel-wise quantization for weights and exponential moving averages (EMA) with momentum to derive the quantization range for activation quantization.

\subsection{Restrict activation ranges}
\label{sec:restrict_activation_range}
\begin{figure*}
\centering
\includegraphics[width=0.45\textwidth, trim=1.5in 0.3in 1.5in 0.8in, clip]{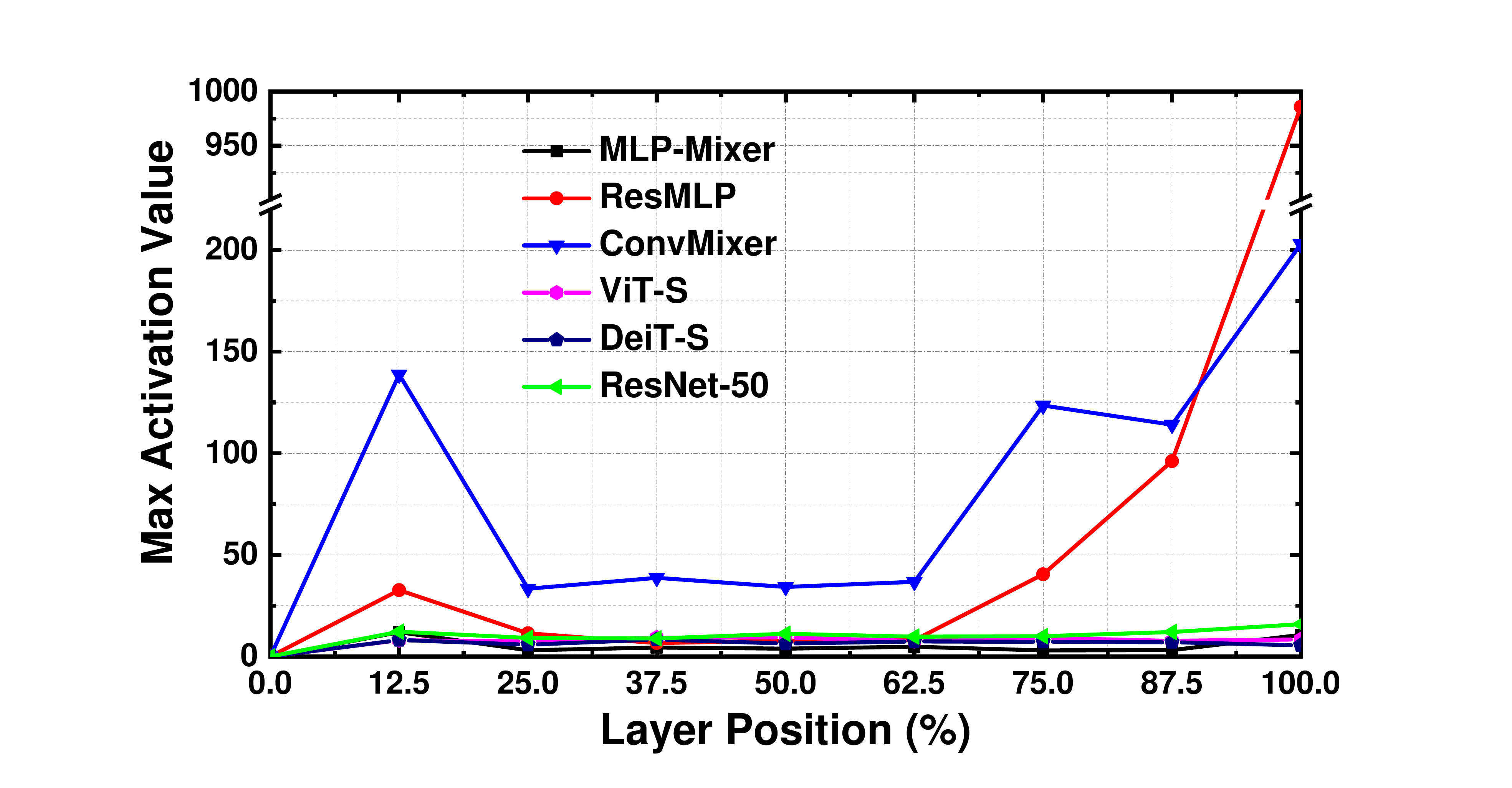}
\includegraphics[width=0.45\textwidth, trim=1.5in 0.3in 1.5in 0.8in, clip]{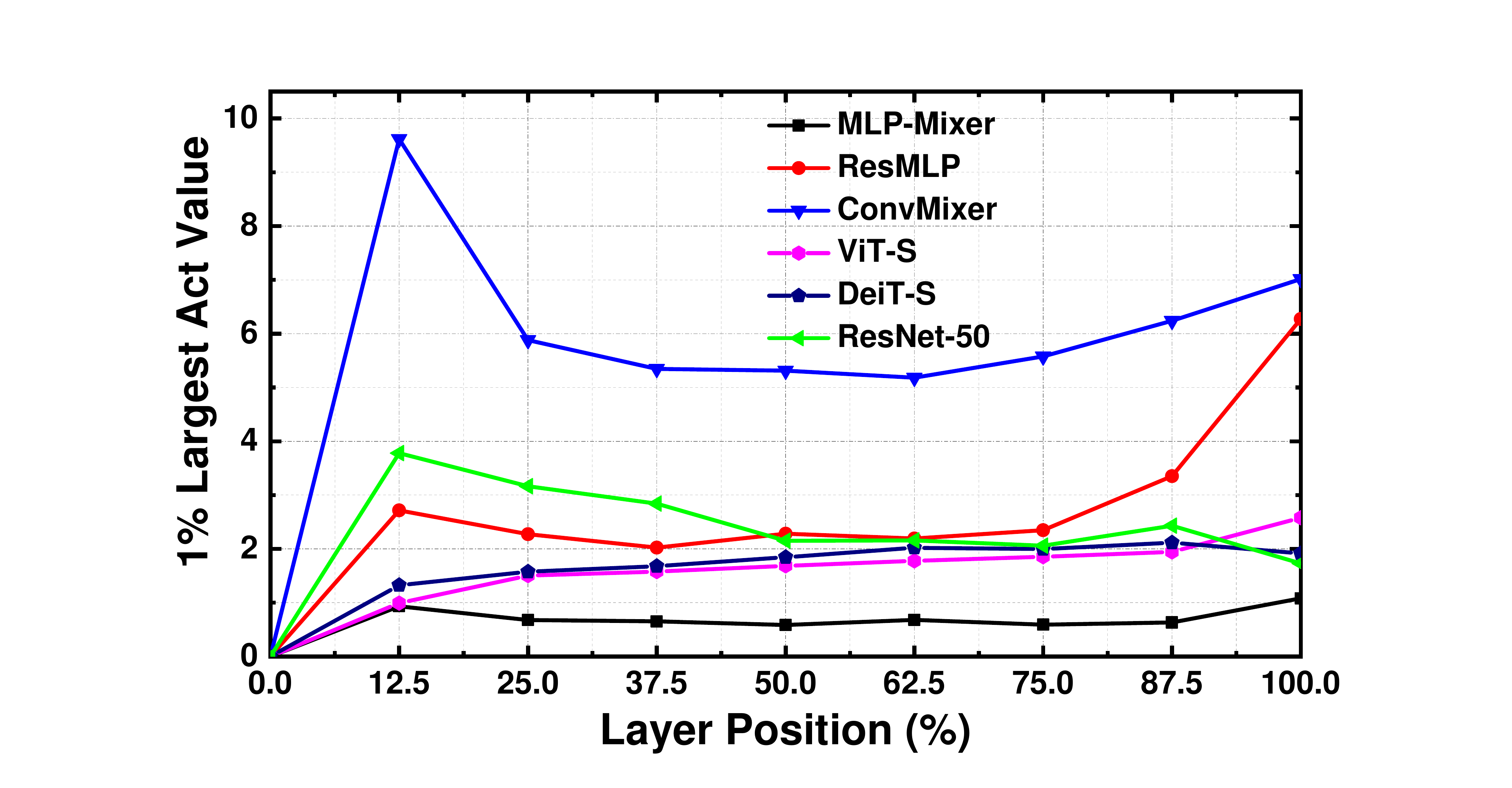}
\includegraphics[width=0.45\textwidth, trim=1.5in 0.3in 1.5in 0.8in, clip]{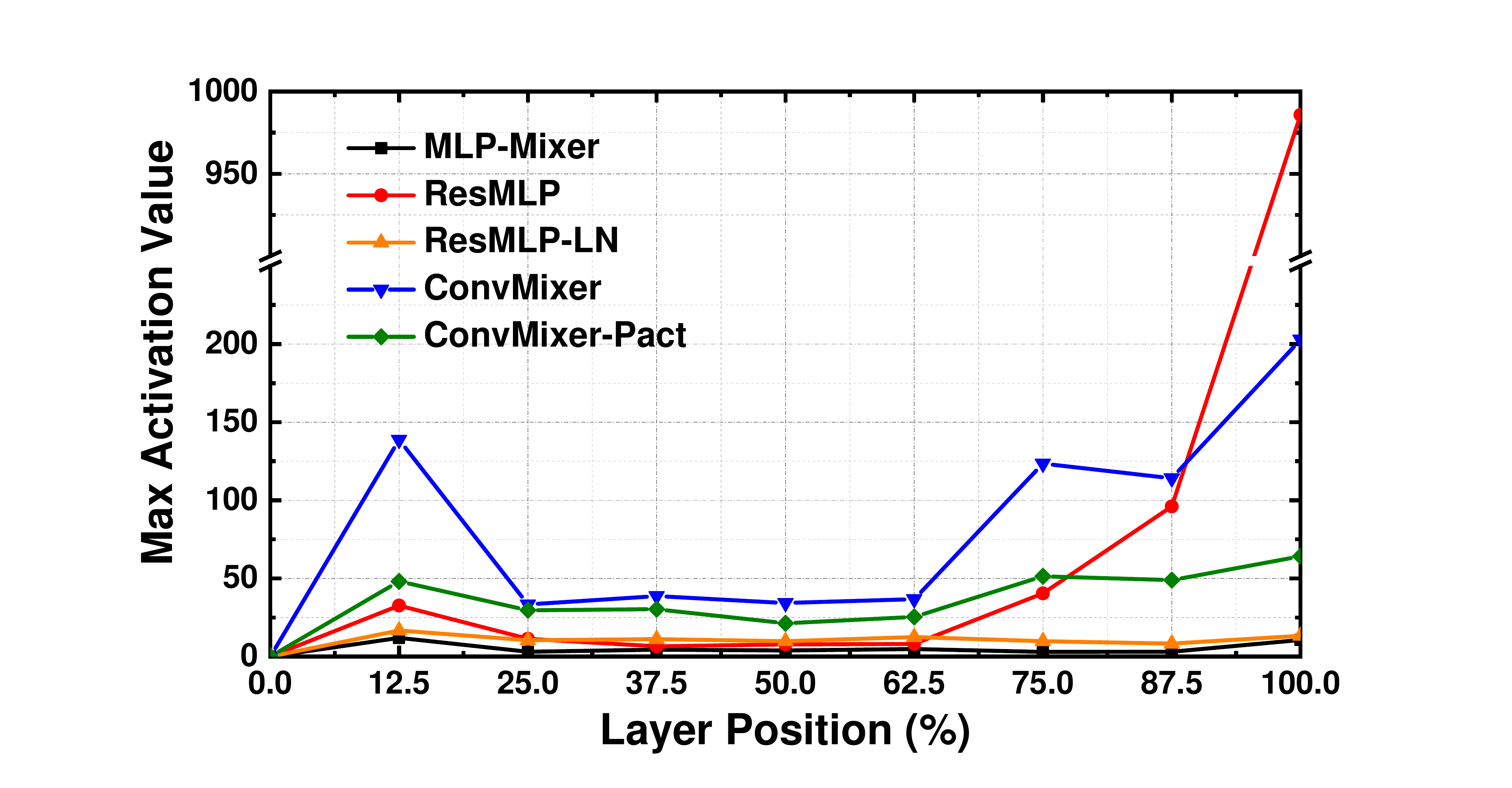}
\includegraphics[width=0.45\textwidth, trim=1.5in 0.3in 1.5in 0.8in, clip]{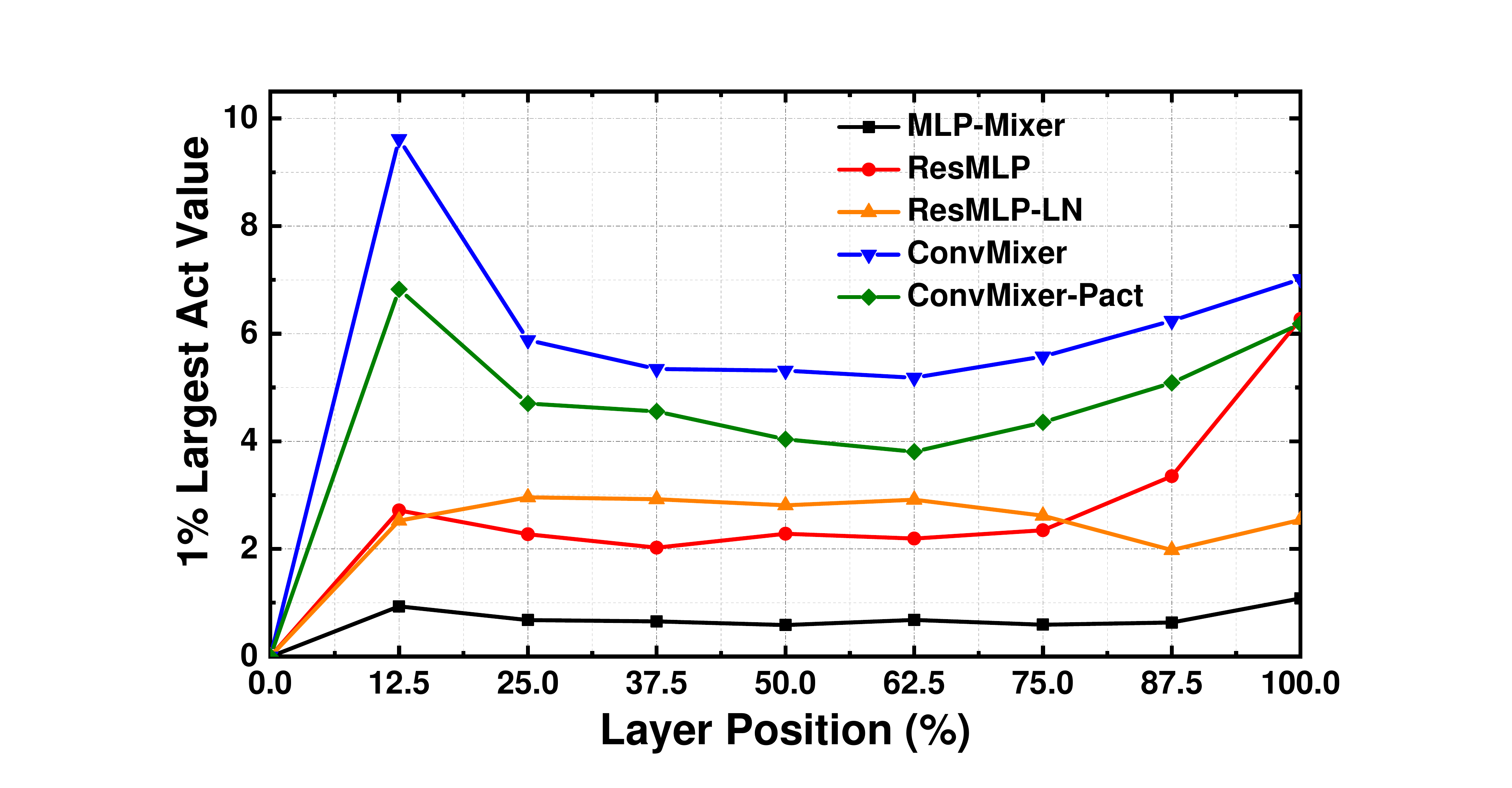}
\caption{
Graph of the max (left) and the 1\% largest (right) activation values in vision models throughout the layers. Layer Position shows the relative position of a given layer in the whole model. For example, the output of the third layer in a 24-layer ResMLP model has a layer position of 12.5\%. The top two graphs show the activation values in MLP, Vision Transformers, and CNNs. The bottom two graphs show activation values in MLPs and our proposed quantization-friendly MLP variants.
}
\label{fig:actvalues_all}
\end{figure*}

In experiments, we find that some MLP-based models have unusually high activation ranges, which lead to large quantization intervals and result in accuracy degradation during quantization. Therefore, we first calculate the max and 99\% percentile of the activation values throughout the layers and plot their activation statistics in Figure \ref{fig:actvalues_all}. These values are crucial for quantization results since they determine the activation range in standard quantization settings and percentile quantization settings, respectively. Results in the top two graphs in Figure \ref{fig:actvalues_all} show that the original ResMLP and ConvMixer models have a relatively high activation range compared to traditional CNNs and Vision Transformers. The bottom two graphs in Figure \ref{fig:actvalues_all} show that after using our proposed methods, such as replacing Affine with LayerNorm in ResMLP and using PACT as activation layers in ConvMixer, the activation range is well restricted. 
In conclusion, these graphs not only present the causes of large accuracy degradation in quantized MLP-based models but also validate the effectiveness of our proposed methods.

\paragraph{Norm-layer design}
As mentioned in Section \ref{sec:method}, quantized MLP-based models may benefit from LayerNorm or BatchNorm. We demonstrate this by replacing the Affine function in ResMLP with LayerNorm. Results in Table \ref{tab:Affine} show that this approach achieves better classification results in PTQ experiments than the original ResMLP model. Moreover, instability issue occurs during QAT experiments for the original ResMLP, and quantization can only be done successfully for ResMLP with LayerNorm. The results suggest that LayerNorm helps restrict the activation range better than the Affine function, which then helps to derive a more accurate integer representation and leads to better accuracy.
\begin{table}[!htbp]
\caption{Affine vs. LayerNorm in ResMLP.
Here, we abbreviate Weight Precision and Activation Precision as ``Precision'',  Norm-layer as ``Norm'', Model Size as ``Size'' (in MB), Bit Operations as ``BOPS'' (in G), and Top-1 Accuracy as ``Top-1''. Note that BOPS is defined as FLOPS $\times$ activation bits $\times $ weight bits, and ``WxAy'' means weight with x-bit and activation with y-bit. Other tables use similar abbreviations in the rest of the paper.
} 
\vspace{5pt}
\label{tab:Affine}
  \begin{center}
    {\small{
\resizebox{0.99\linewidth}{!}{
\setlength\tabcolsep{4 pt}
\begin{tabular}{cccccc} \toprule
\multicolumn{1}{c}{\textbf{Method}} 
&\multicolumn{1}{c}{\textbf{Precisioin }} 
&\multicolumn{1}{c}{\textbf{Norm} }
&\multicolumn{1}{c}{\textbf{Size(MB)} }
&\multicolumn{1}{c}{\textbf{BOPS(G)} }
&\multicolumn{1}{c}{\textbf{Top-1} } \\
\midrule

\ha \multirow{2}{*}{Baseline}  & \multirow{2}{*}{W32A32}   
        &  Affine   & 120  & 12226   &  79.38   \\
&       &  LN       & 120  & 12226   &  79.59   \\
\midrule
\ha \multirow{2}{*}{PTQ}   &\multirow{2}{*}{W8A8 }     
        &  Affine   & 30   &764      &   74.93  \\
&       &  LN       & 30   & 764     &  79.20   \\
\midrule
\ha \multirow{2}{*}{QAT}   &\multirow{2}{*}{W8A8 }         
        &  Affine   & 30   & 764     & -        \\
&       &  LN       & 30   & 764     &  79.44   \\
\midrule
\ha   QAT    
& W4A8  &  LN       & 15   &382      &  78.35   \\
\bottomrule 
\end{tabular}
}}
}
\end{center}
\end{table}

\paragraph{Activation layer design}
For models with an extremely large activation range, using LayerNorm or BatchNorm may not be sufficient for restricting the activation range. For example, ConvMixer, using BatchNorm as its norm layer, still suffers from quantization degradation due to its large activation range. However, results in Table \ref{tab:PACT} demonstrate that we can efficiently restrict the activation range by using activation layers with bounded outputs for both negative and positive input values. With the help of a narrower activation range, our PACT-ConvMixer achieves at least similar results in different PTQ settings and much better results in all settings of QAT. Here, we choose to use PACT activation in our quantized ConvMixer model since it has a learnable upper bound and is potentially more capable in restricting the range. Other bounded activation layers (for example, ReLU6) should work as well. It is important to mention that we use the asymmetric quantization settings for QAT comparison since QAT for ConvMixers cannot converge in symmetric settings. A potential reason is that QAT suffers more from sensitive activation ranges, and asymmetric quantization reduces activation sensitivity, which will be discussed in detail in Section \ref{sec:epsensitive}.

\begin{table}[!htbp]
\caption{ReLU vs. PACT in ConvMixer} 
\vspace{5pt}
\label{tab:PACT}
  \begin{center}
    {\small{
\resizebox{0.99\linewidth}{!}{
\setlength\tabcolsep{4 pt}
\begin{tabular}{cccccc} \toprule
\multicolumn{1}{c}{\textbf{Method}} 
&\multicolumn{1}{c}{\textbf{Precisioin }} 
&\multicolumn{1}{c}{\textbf{Activation} }
&\multicolumn{1}{c}{\textbf{Size(MB)} }
&\multicolumn{1}{c}{\textbf{BOPS(G)} }
&\multicolumn{1}{c}{\textbf{Top-1} } \\
\midrule 
\ha \multirow{2}{*}{Baseline}  & \multirow{2}{*}{W32A32  }   
        & ReLU    &  84     & 42762     &  80.16     \\
 &      & PACT    &  84     & 42762     &  80.22     \\
\midrule
\ha \multirow{2}{*}{PTQ}  & \multirow{2}{*}{W8A8}          
        & ReLU   & 21      & 2672       &  57.81    \\
&       & PACT   & 21      & 2672       &  68.91    \\
\midrule
\ha \multirow{2}{*}{QAT(asym) }  & \multirow{2}{*}{W8A8}    
        & ReLU   & 21      & 2672       & 77.09     \\
&       & PACT   & 21      & 2672       & 78.65     \\
\midrule
\ha \multirow{2}{*}{QAT(asym)} & \multirow{2}{*}{W4A8}  
        & ReLU   & 11      & 1336       & 75.88     \\
&       & PACT   & 11      & 1336       & 77.89     \\
\bottomrule 
\end{tabular}
}}
}
\end{center}
\end{table}

\paragraph{Percentile}
An easier way to restrict the activation range is to use a percentile max value when calculating the quantization parameters. For example, using a 99\% percentile option can help clip the 1\% biggest activation values so that the activation range will no longer depend on those extreme outliers. Table \ref{tab:Percentile} shows that percentile partly helps to recover the accuracy of the quantized models.
Note that calculating percentiles in QAT make the training process much slower, so we only use activation percentiles in PTQ experiments.

\begin{table}[!htbp]
\caption{Percentile in ResMLP} 
\vspace{5pt}
\label{tab:Percentile}
  \begin{center}
    {\small{
\resizebox{0.99\linewidth}{!}{
\setlength\tabcolsep{4 pt}
\begin{tabular}{cccccc} \toprule
\multicolumn{1}{c}{\textbf{Method}} 
&\multicolumn{1}{c}{\textbf{Precisioin }} 
&\multicolumn{1}{c}{\textbf{Percentile} }
&\multicolumn{1}{c}{\textbf{Size(MB)} }
&\multicolumn{1}{c}{\textbf{BOPS(G)} }
&\multicolumn{1}{c}{\textbf{Top-1} } \\
\midrule
\ha Baseline             & W32A32     
        & $\times$         & 120   & 12226 &   79.38   \\
\midrule
\ha \multirow{2}{*}{PTQ}    & \multirow{2}{*}{W8A8}      
        & $\times$          & 30    & 764   &   74.93   \\
&       & $\checkmark$      & 30    & 764   &   77.74   \\
\bottomrule
\end{tabular}
}}
}
\end{center}
\end{table}

\subsection{Tackle sensitive layers}\label{sec:epsensitive}
\begin{figure}[ht]
\begin{center}

\includegraphics[width=0.9\linewidth, trim=1.5in 0.3in 1.5in 0.8in, clip]{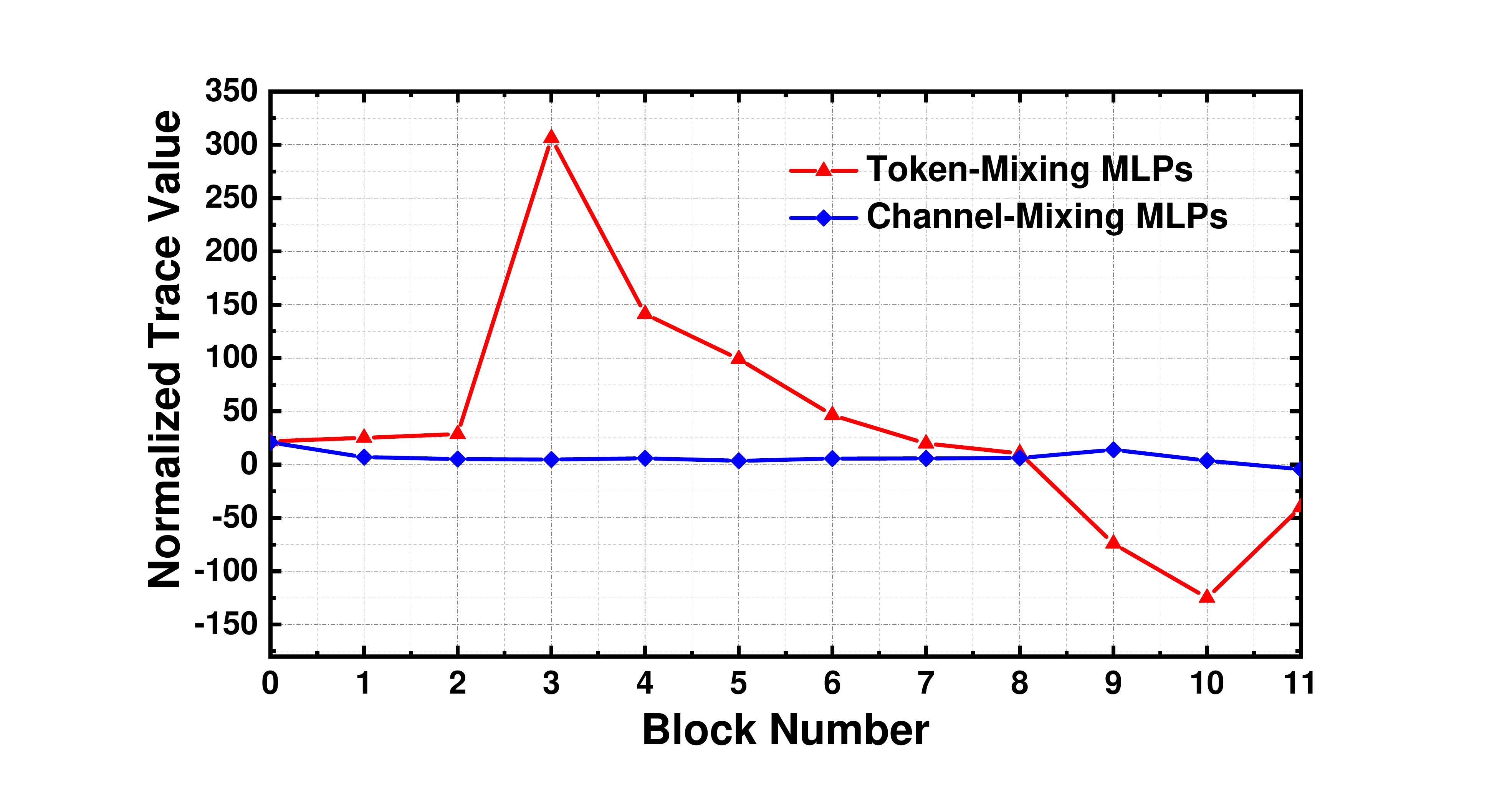}
\caption{Mean Hessian traces of token-mixing MLP and channel-mixing MLP in layers of MLP-Mixer. Note that the Hessian trace values are normalized according to their parameter size.}
  \label{fig:trace}
    
\end{center}
\end{figure}

\paragraph{Multiple token-mixing MLPs}
As mentioned in Section \ref{sec:sensitive}, we calculate the Hessian traces of each MLP in MLP-Mixer in Figure \ref{fig:trace}. Results show that parameters in token-mixing MLPs are more sensitive than in channel-mixing MLPs.  Therefore, token-mixing MLPs should be carefully designed in order to achieve better PTQ results.

In Table \ref{tab:Sensitivity} we show that the full precision accuracy, PTQ, and QAT results all obtain a remarkable improvement after introducing the multiple token-mixing MLP into the original MLP model. It indicates that reducing the sensitivity of specific parameters is crucial for obtaining high-performance quantized MLP-based models.

\begin{table}[!htbp]
\caption{Token-mixing in MLP-Mixer} 
\vspace{5pt}
\label{tab:Sensitivity}
  \begin{center}
    {\small{
\resizebox{0.99\linewidth}{!}{
\setlength\tabcolsep{4 pt}
\begin{tabular}{cccccc} \toprule
\multicolumn{1}{c}{\textbf{Method}} 
&\multicolumn{1}{c}{\textbf{Precisioin }} 
&\multicolumn{1}{c}{\textbf{Token-mixing} }
&\multicolumn{1}{c}{\textbf{Size(MB)} }
&\multicolumn{1}{c}{\textbf{BOPS(G)} }
&\multicolumn{1}{c}{\textbf{Top-1} } \\
\midrule
\ha \multirow{2}{*}{Baseline}    & \multirow{2}{*}{W32A32}             
        &  Single      & 240  &25825  &   76.64   \\
&       &  Multiple    & 261  &25825  &   78.35   \\
\midrule
\ha \multirow{2}{*}{PTQ}    & \multirow{2}{*}{W8A8 }          
        &  Single      & 60   &1614   &   74.46  \\
&       &  Multiple    & 65   &1614   &   75.34 \\
\midrule
\ha \multirow{2}{*}{QAT}    & \multirow{2}{*}{W4A8 }         
        &  Single      & 30   &807    &   75.82  \\
&       &  Multiple    & 33   &807    &   77.66 \\
\midrule
\ha QAT       & W3A8       &  Multiple   & 25  &605  &  76.85 \\
\bottomrule 
\end{tabular}
}}
}
\end{center}
\end{table}

\paragraph{Asymmetric quantization}
As described in Section \ref{sec:sensitive}, asymmetric quantization can help ease the sensitivity of activation layers by adding an extra bitwidth implicitly. Table \ref{tab:Asymmetric} shows that asymmetric quantization is helpful in quantized MLP-based models and especially important in the quantization of ConvMixer. We also find that ConvMixer can only use QAT in the asymmetric quantization mode, and the training would be very likely to diverge otherwise. These results imply that using asymmetric quantization to reduce sensitivity not only provides better performance but also helps stabilize QAT.

\begin{table}[!htbp]
\caption{Symmetric vs. Asymmetric in PTQ experiments} 
\vspace{5pt}
\label{tab:Asymmetric}
\centering
    {\small{
\resizebox{0.99\linewidth}{!}{
\setlength\tabcolsep{2 pt}
\begin{tabular}{lccccc} \toprule
\multicolumn{1}{c}{\textbf{Model}} 
&\multicolumn{1}{c}{\textbf{Precisioin }} 
&\multicolumn{1}{c}{\textbf{Sym/Asym} }
&\multicolumn{1}{c}{\textbf{Size(MB)} }
&\multicolumn{1}{c}{\textbf{BOPS(G)} }
&\multicolumn{1}{c}{\textbf{Top-1} } \\
\midrule
\ha \multirow{2}{*}{Q-MLP-Mixer}    & \multirow{2}{*}{W8A8}        
        &  Sym       & 60  & 1614   &74.46   \\
&       &  Asym      & 60  & 1614   &76.20   \\
\midrule
\ha \multirow{2}{*}{Q-ResMLP}    & \multirow{2}{*}{W8A8}           
        &  Sym       & 30   & 764 & 74.93   \\
&       &  Asym      & 30   & 764 & 78.28   \\
\midrule
\ha \multirow{2}{*}{Q-ConvMixer}    & \multirow{2}{*}{W8A8}      
        &  Sym       & 21  & 2672 & 57.81    \\
&       &  Asym      & 21  & 2672 & 76.21    \\
\bottomrule 
\end{tabular}
}}
}
\end{table}

\subsection{Ablation study}
Here, we take ConvMixer as an example and combine all the techniques mentioned above to provide a thorough ablation study to illustrate the effectiveness of our methods. As the results shown in Table~\ref{tab:Ablation}, bounded activation layer, activation percentiles, and asymmetric quantization not only improve the quantization performance separately, but they can boost the performance with any of the combinations. Incorporating all of the aforementioned methods gives us the best results for ConvMixer with an 8-bit PTQ of 76.78\% ImageNet classification accuracy.

\begin{table}[!htbp]
\caption{Ablation Study} 
\vspace{5pt}
\label{tab:Ablation}
  \begin{center}
    {\small{
\resizebox{0.99\linewidth}{!}{
\setlength\tabcolsep{0.8 pt}
\begin{tabular}{cccccc} \toprule
\multicolumn{1}{c}{\textbf{Model}} 
&\multicolumn{1}{c}{\textbf{BatchNorm}} 
&\multicolumn{1}{c}{\textbf{Percentile} }
&\multicolumn{1}{c}{\textbf{Asymmetric} }
&\multicolumn{1}{c}{\textbf{PACT} }
&\multicolumn{1}{c}{\textbf{Top-1} } \\
\midrule
ConvMixer &- &- &- &- &80.16 \\
\midrule
\ha \multirow{8}{*}{Q-ConvMixer}      
    & $\checkmark$    &  $\times$       & $\times$      & $\times$      &  57.81    \\
    & $\checkmark$    &  $\checkmark$   & $\times$      & $\times$      &  69.36    \\
    & $\checkmark$    &  $\times$       & $\checkmark$  & $\times$      &  76.21    \\
    & $\checkmark$    &  $\checkmark$   & $\checkmark$  & $\times$      &  76.35    \\
    & $\checkmark$    &  $\times$       & $\times$      & $\checkmark$  &  68.91    \\
    & $\checkmark$    &  $\checkmark$   & $\times$      & $\checkmark$  &  73.88    \\
    & $\checkmark$    &  $\times$       & $\checkmark$  & $\checkmark$  &  76.06    \\
    & $\checkmark$    &  $\checkmark$   & $\checkmark$  & $\checkmark$  &  76.78    \\
\bottomrule 
\end{tabular}
}}
}
\end{center}
\end{table}

\subsection{Best quantized models}
Combining all of the abovementioned methods, we derive the best results for quantized MLP-Mixer, ResMLP, ConvMixer in Table \ref{tab:PTQcompare} and \ref{tab:QATcompare} with boldface.
Since we are the first to investigate quantization aspects of MLP models, there are few previous works to compare with. Therefore, we apply the open-sourced CNN-targeted quantization method HAWQ-v3 \cite{yao2021hawq}, which supports both PTQ and QAT, to MLP-based models for comparison. Results in Table \ref{tab:PTQcompare} and \ref{tab:QATcompare} show that our quantization method works much better on MLP-based models, indicating the importance of considering the MLP models' particular structure. Although all three models gain much better accuracy after applying our proposed methods, ResMLP distinctly outperforms the other two MLP-based models in experiments. It implies that ResMLP variants are potentially more efficient and quantization friendly. 
Meanwhile, though MLP-Mixer's performance and computation-accuracy trade-off are slightly behind the other two models, it is the easiest to quantize among the three MLP-based models. QAT on MLP-Mixer can be conducted smoothly, while instability issue occurs in ResMLP and ConvMixer unless we redesign the norm-layer and activation layer according to Section \ref{sec:method}. 

To compare our best quantized MLP-based models with quantized CNNs and transformer-based networks, we highlight the best quantized MLP-based models with different precision settings in Table \ref{tab:PTQcompare} and \ref{tab:QATcompare}. Results show that Q-ResMLP outperforms other quantized models with similar model scales and can even achieve comparable performance with some much larger models.

\begin{table*}
\caption{Comparison of the post-training quantization performance of MLP-based models with CNNs and transformer-based models using different quantization methods.}
\vspace{5pt}
\label{tab:PTQcompare}
  \begin{center}
    {\small{
\resizebox{0.90\linewidth}{!}{
\setlength\tabcolsep{6 pt}
\begin{tabular}{l|lccccc} \toprule
\multicolumn{1}{c}{\textbf{Category}}
&\multicolumn{1}{c}{\textbf{Model}} 
&\multicolumn{1}{c}{\textbf{Method} }
&\multicolumn{1}{c}{\textbf{Precisioin }} 
&\multicolumn{1}{c}{\textbf{Size(MB)} }
&\multicolumn{1}{c}{\textbf{BOPS(G)} }
&\multicolumn{1}{c}{\textbf{Top-1} } \\
\midrule
\multirow{6}{*}{MLP-based Networks} 
&\multirow{2}{*}{Q-MLP-Mixer}  
    &HAWQ-V3 \cite{yao2021hawq}  &W8A8   & 60    &1614   & 74.40   \\
&   &Ours                        &W8A8   & 65    &1614   & \textbf{77.75}   \\
\cmidrule{2-7}
&\multirow{2}{*}{Q-ResMLP}
    &HAWQ-V3 \cite{yao2021hawq}  &W8A8   & 30    &764    & 76.69   \\
&   &\cellcolor{orange!40}Ours &\cellcolor{orange!40}W8A8     &\cellcolor{orange!40}30    &\cellcolor{orange!40}764    &\cellcolor{orange!40}\textbf{79.43}   \\
\cmidrule{2-7}
 &\multirow{2}{*}{Q-ConvMixer}
    &HAWQ-V3 \cite{yao2021hawq}  &W8A8   & 21    &2672   & 72.54   \\
&   &Ours                        &W8A8   & 21    &2672   & \textbf{76.78}   \\
\midrule
\multirow{6}{*}{Transformer Networks} 
&\multirow{2}{*}{Q-DeiT-S}    
    &EasyQuant \cite{wu2020easyquant} &W8A8   & 22    &543   & 76.59   \\
&   &Bit-Split \cite{wang2020towards} &W8A8   & 22    &543   & 77.06   \\
\cmidrule{2-7}
&\multirow{2}{*}{Q-DeiT-B}    
 &EasyQuant \cite{wu2020easyquant}  &W8A8    & 86    &2158   & 79.36   \\
& &Bit-Split \cite{wang2020towards}  &W8A8    & 86    &2158   & 79.42   \\
\cmidrule{2-7}
&\multirow{2}{*}{Q-ViT-B}    
  &Percentile \cite{li2019fully}    &W8A8   & 86    &2158   & 74.10   \\
& &PTQ-ViT    \cite{liu2021post}    &W8A8   & 86    &2158   & 76.98   \\
\midrule
\multirow{2}{*}{Convolutional Networks} 
&\multirow{2}{*}{Q-ResNet50}    
  &Bit-Split  \cite{wang2020towards} &W8A8  & 26    &496    & 75.96   \\
& &ZeroQ \cite{cai2020zeroq}  &W8A8    & 26    &496   & 77.67   \\
\bottomrule 
\end{tabular}
}}
}
\end{center}
\end{table*}

\begin{table*}
\caption{Comparison of the quantization-aware training performance of MLP-based models with CNNs using different quantization methods (we did not find QAT results of transformers in previous works). Here, 4/8 in HAWQ-V3 means mixed precision with 4 and 8 bits. Note that Q-ResMLP and Q-ConvMixer results using HAWQ-V3 are derived after applying our norm-layer and activation-layer design to the original model.}
\vspace{5pt}
\label{tab:QATcompare}
  \begin{center}
    {\small{
\resizebox{0.90\linewidth}{!}{
\setlength\tabcolsep{6 pt}
\begin{tabular}{l|lccccc} \toprule
\multicolumn{1}{c}{\textbf{Category}}
&\multicolumn{1}{c}{\textbf{Model}} 
&\multicolumn{1}{c}{\textbf{Method} }
&\multicolumn{1}{c}{\textbf{Precisioin }} 
&\multicolumn{1}{c}{\textbf{Size(MB)} }
&\multicolumn{1}{c}{\textbf{BOPS(G)} }
&\multicolumn{1}{c}{\textbf{Top-1} } \\
\midrule
\multirow{10}{*}{MLP-based Networks} 
&\multirow{3}{*}{Q-MLP-Mixer}  
    &HAWQ-V3 \cite{yao2021hawq}  &W8A8   & 60    &1614   & 76.28   \\
&   &Ours                        &W8A8   & 65    &1614   & \textbf{78.17}   \\
&   &Ours                        &W4A8   & 33    &807   & \textbf{77.94}   \\
\cmidrule{2-7}
&\multirow{3}{*}{Q-ResMLP}
    &HAWQ-V3 \cite{yao2021hawq}  &W8A8   & 30    &764    & 77.36   \\
&   &\cellcolor{orange!40}Ours &\cellcolor{orange!40}W8A8     &\cellcolor{orange!40}30    &\cellcolor{orange!40}764    &\cellcolor{orange!40}\textbf{79.68}   \\
&   &\cellcolor{orange!40}Ours &\cellcolor{orange!40}W4A8     &\cellcolor{orange!40}15    &\cellcolor{orange!40}382    &\cellcolor{orange!40}\textbf{78.47}   \\
\cmidrule{2-7}
 &\multirow{3}{*}{Q-ConvMixer}
    &HAWQ-V3 \cite{yao2021hawq}  &W8A8   & 21    &2672   & 75.88   \\
&   &Ours                        &W8A8   & 21    &2672   & \textbf{78.65}   \\
&   &Ours                        &W4A8   & 11    &1336   & \textbf{77.89}   \\
\midrule
\multirow{5}{*}{Convolutional Networks} 
&\multirow{5}{*}{Q-ResNet50}    
&Integer Only \cite{jacob2018quantization} &W8A8  & 26    &496   & 74.90   \\
& &RVQuant \cite{park2018value} &W8A8  & 26    &496   & 75.67   \\
& &HAWQ-V3 \cite{yao2021hawq} &W8A8  & 26    &496   & 77.58   \\
& &HAWQ-V3 \cite{yao2021hawq} &W4/8A4/8  & 19    &308   & 75.39   \\
& &LQ-Nets \cite{zhang_2018_lqnets} &W4A32  & 13    &992   & 76.40   \\
\bottomrule 
\end{tabular}
}}
}
\end{center}
\end{table*}

%% file: s6_conclusion.tex
\section{Conclusions}\label{sec:conclusion}
\vspace{-3pt}
In this work, we analyze the quantization of state-of-the-art MLP-based vision models. Two major problems concluded are: 1) MLP-based models suffer from large quantization ranges of activations 2) specific components of MLP-based models are sensitive to quantization. To alleviate the degradation caused by activations, we propose to apply LayerNorm to control the quantization range, and we also take advantage of bounded activation functions and percentile quantization on activations. In order to tackle the sensitivity, we propose to apply our multiple token-mixing MLPs and use linear asymmetric quantizers for the sensitive operations in MLP-based models. With these practical techniques, our Q-MLP models can achieve 79.43\% top-1 accuracy on ImageNet with 8-bit post-training quantization (30 MB model size). For quantization-aware training, our Q-MLP models can achieve 79.68\% accuracy using 8-bit (30 MB) and 78.47\% accuracy using 4-bit quantization (15 MB).

%% file: s8_supp.tex
\section{Implementation details}
\label{sec:detail}
We primarily evaluate our proposed and existing models on the ImageNet-1k validation set. Specifically, we add our Q-MLP-Mixer, Q-ResMLP, Q-ConvMixer models into the timm framework \cite{timm}, and then train new models and implement the Quantization-Aware Traning (QAT) under the default settings in \cite{tolstikhin2021mlp,touvron2021resmlp,trockman2022patches} except changing the initial learning rate to $2\times10^{-5}$ during QAT. The training for new models, such as multiple token-mixing MLPs and the ResMLP with LayerNorm, usually takes 4-5 days to train on eight TITAN RTX 2080Ti (24GB) GPUs, and the QAT experiments usually take 1-2 days.

It should be noted that the three papers~\cite{tolstikhin2021mlp,touvron2021resmlp,trockman2022patches} mentioned above provide several variants for each model architecture. We choose Mixer-B/16 for MLP-Mixer, ResMLP-S24 for ResMLP, and ConvMixer-768/32 for ConvMixer. These three models are relatively similar in the number of parameters so that we can make reasonable comparisons.

\section{Additional results}
This section presents some additional results that are not mentioned in the main paper. These results either do not help much to derive the main conclusions or happen to be unsuccessful attempts. However, these results may still be meaningful for future investigation.

\subsection{4-bit post-training quantization (PTQ) results}
4-bit PTQ results are not mentioned in our main contents because 4-bit quantization settings usually require QAT to achieve desirable performance. However, some of these 4-bit PTQ results may bring insights to the MLP-based models. We include the 4bit PTQ results of baselines and the best quantization model of each MLP variant in Table \ref{tab:PTQ4bit}. We should emphasize that we did not use percentile quantization to improve the accuracy here since we want to reflect each model's potential in 4-bit PTQ directly. However, we use asymmetric quantization in all 4-bit PTQ experiments to ease the activation sensitivity issue. Results show that MLP-Mixer does not suffer much under 4-bit quantization settings, while ResMLP and ConvMixer encounter severe accuracy degradation. These results are consistent with our analysis in the main paper that models with large activation ranges suffer more in quantization, and it also implies that the original MLP-Mixer has potential in ultra-low-bit quantization due to its uniform structure.

\begin{table}[!htbp]
\caption{PTQ results of 4-bit quantization. 
} 
\vspace{5pt}
\label{tab:PTQ4bit}
  \begin{center}
    {\small{
\resizebox{0.99\linewidth}{!}{
\setlength\tabcolsep{1 pt}
\begin{tabular}{cccccc} \toprule
\multicolumn{1}{c}{\textbf{Model}} 
&\multicolumn{1}{c}{\textbf{Precisioin }} 
&\multicolumn{1}{c}{\textbf{Method} }
&\multicolumn{1}{c}{\textbf{Size(MB)} }
&\multicolumn{1}{c}{\textbf{BOPS(G)} }
&\multicolumn{1}{c}{\textbf{Top-1} } \\
\midrule 
\ha \multirow{2}{*}{MLP-Mixer}  & \multirow{2}{*}{W32A32  }   
        & Token-mixing          &  240     & 25825     &  76.64     \\
 &      & Multi-token-mixing    &  261     & 25825     &  78.35     \\
\midrule 
\ha \multirow{2}{*}{Q-MLP-Mixer}  & \multirow{2}{*}{W4A8  }   
        & Token-mixing          &  30     & 807     &  75.82     \\
 &      & Multi-token-mixing    &  33     & 807     &  76.99     \\
\midrule 
\ha \multirow{2}{*}{ResMLP}  & \multirow{2}{*}{W32A32  }   
        & Affine    &  120     & 12226     &  79.38     \\
 &      & LN        &  120     & 12226     &  79.59     \\
\midrule 
\ha \multirow{2}{*}{Q-ResMLP}  & \multirow{2}{*}{W4A8  }   
        & Affine    &  15     & 382     &  60.67     \\
 &      & LN        &  15     & 382     &  61.12     \\
\midrule 
\ha \multirow{2}{*}{ConvMixer}  & \multirow{2}{*}{W32A32  }   
        & ReLU    &  84     & 42762     &  80.16     \\
 &      & PACT    &  84     & 42762     &  80.22     \\
\midrule 
\ha \multirow{2}{*}{Q-ConvMixer}  & \multirow{2}{*}{W4A8  }   
        & ReLU    &  11     & 1336     &  60.67     \\
 &      & PACT    &  11     & 1336     &  63.73     \\
\bottomrule 
\end{tabular}
}}
}
\end{center}
\end{table}

\subsection{GELU vs. ReLU}
We found that replacing GELU with ReLU does not help to improve the quantization performance or restrict the activation range. As shown in Table \ref{tab:gelurelu}, GELU seems better for ResMLP, while ReLU works better for ConvMixer. However, the difference between GELU and ReLU is negligible, and small randomness during the training process may cause the difference between the two variants. The activation ranges of GELU and ReLU are also similar since the max absolute values of the two activations are close. The noticeable accuracy degradation in Table~\ref{tab:gelurelu} implies that we need bounded activation functions (e.g., PACT) to deal with extremely large activation ranges, as discussed in the main text.

\begin{table}[!htbp]
\caption{GELU vs. ReLU} 
\vspace{5pt}
\label{tab:gelurelu}
  \begin{center}
    {\small{
\resizebox{0.99\linewidth}{!}{
\setlength\tabcolsep{2 pt}
\begin{tabular}{cccccc} \toprule
\multicolumn{1}{c}{\textbf{Method}} 
&\multicolumn{1}{c}{\textbf{Precisioin }} 
&\multicolumn{1}{c}{\textbf{Activation} }
&\multicolumn{1}{c}{\textbf{Size(MB)} }
&\multicolumn{1}{c}{\textbf{BOPS(G)} }
&\multicolumn{1}{c}{\textbf{Top-1} } \\
\midrule 
\ha \multirow{2}{*}{ResMLP}  & \multirow{2}{*}{W32A32}   
        & GELU    &  120     & 12226     &  79.38     \\
 &      & ReLU    &  120     & 12226     &  79.19     \\
\midrule
\ha \multirow{2}{*}{Q-ResMLP}  & \multirow{2}{*}{W8A8}          
        & GELU    &  30     & 764     &  79.20     \\
 &      & ReLU    &  30     & 764     &  78.52     \\
\midrule
\ha \multirow{2}{*}{ConvMixer}  & \multirow{2}{*}{W32A32}    
        & GELU    &  84     & 42762     &  79.73     \\
 &      & ReLU    &  84     & 42762     &  80.16     \\
\midrule
\ha \multirow{2}{*}{Q-ConvMixer} & \multirow{2}{*}{W4A8}  
        & GELU    &  11     & 2672       & 52.39     \\
&       & ReLU    &  11     & 2672       & 57.81     \\
\bottomrule 
\end{tabular}
}}
}
\end{center}
\end{table}

\subsection{Restricting activation range}
ConvMixer has an extremely large activation range and faces severe performance degradation. In contrast to our methods proposed in the main contents, we would also like to discuss a few other methods that fail to restrict the activation range. 

Firstly, exchanging the position of BNs and convolutions in each ConvMixer layer does not help to restrict the activation range. Besides, including a weight-decay term during training is also not helpful. Though the weight-decay option may slightly restrict the weight values, the activation range remains almost the same. 

\section{Limitation and future work}
Though we have presented many tables and ablation studies in Section 4 in the main paper, some work remains to be done in the future. Firstly, since big model variants with different structures have relatively large differences in terms of parameter size and FLOPs, it is hard for us to make fair comparisons to all of them. Consequently, our experimental results mainly focus on a few selected (relatively small) MLP variants, and more results among larger MLPs, CNNs, and transformer models remain to be explored in future work. Secondly, we use models pretrained solely on ImageNet. In \cite{tolstikhin2021mlp}, the authors state that MLP-Mixer tends to overfit more than ViT, which implies that MLP-Mixer will potentially benefit more when pre-trained with larger datasets (for example, JFT-300M). Therefore, it is interesting to include some model variants pretrained on larger datasets to see if more data benefit MLP-based models more than transformers and CNNs in their quantization results.
Lastly, our work only explores the merit of uniform quantization in order to maximize efficiency during inference. Aspects of mixed-precision quantization can be explored in future work.